Marica Notte (Institute of Cognitive Sciences and Technologies (ISTC-CNR)),
Ludovica Marinucci (Interdepartmental Center for Research Ethics and Integrity (CID Ethics), Vieri Giuliano Santucci (ISTC-CNR)

# Autonomy, Social Norms, and Alignment: Towards a Developmental Framework for Autonomous Artificial Agents

## 1. *Autonomy in artificial embodied agents*

In recent years, artificial intelligence (AI) has grown massively in its role and impact on daily-life activities. Large-model-based systems, both text-and visual-based, now constitute established tools supporting numerous human activities, especially in knowledge-related work (Brachman et al., 2025). Alongside, robotics has significantly improved its capability to develop embodied agents capable of real-time interaction, complex action sequences, and forms of social interaction, leading to envisage them as increasingly expected to operate as collaborators rather than programmable tools. Despite this progress, the majority of current AI systems (both software and embodied) rely on massive pretraining over large datasets and/or continuous human feedback: although powerful, such strategies may prove insufficient in dynamic, non-structured, and potentially unknown real-world scenarios, where artificial agents need to acquire knowledge and learn new skills through interaction in order to adapt and fulfil human requests (Silver & Sutton, 2025).

To address the challenge of increasing adaptiveness and versatility in artificial agents, and in particular in embodied ones, machine learning techniques have been developed within the field of developmental robotics (Cangelosi & Schlesinger, 2015) and the framework of intrinsically motivated open-ended learning (Santucci et al., 2020). Drawing inspiration from concepts such as curiosity, competence acquisition, and empowerment, these techniques can guide autonomous artificial agents (AAA) in the exploration of complex environments, in the discovery of new goals and the learning of related skills, as well as in the construction of curricula for solving complex tasks and adapting to non-stationary environments (Colas et al., 2022; Romero et al., 2025). Such techniques exemplify how, to face complex real-world scenarios, robots must be endowed with higher levels of autonomy than simple training over predefined datasets: an autonomy intended as the capability to set their own goals driven by the broad motivation of expanding the knowledge of the system, to prioritise learning on the basis of self-generated signals, and to gather new information through curiosity-driven exploration. While such learning autonomy improves robots' adaptiveness and capacity to solve assigned tasks (Mendonca et al., 2021), the expansion of their decisional autonomy makes it more complex to guarantee that such autonomous agents remain aligned with the goals and preferences of designers and users, as well as with social rules and norms.

In this context, the alignment problem (Christian, 2020), i.e. the challenge of ensuring that AI systems behave in accordance with human values, ethics, and intended goals, already difficult for AI in general, becomes even more complex for AAA operating in unstructured and possibly unknown scenarios, where it is not feasible to anticipate all events and where pre-programmed sets of rules might prove insufficient, or might dampen the very autonomy the robots need. While a top-down, normative approach is viable in controlled environments and with low-autonomy agents (Dyoub et al., 2026), it is not for high-autonomy ones. Therefore, to hold together the need for autonomy and the requirement of alignment with human values, we propose to consider the alignment problem as an epistemic challenge, in which norms, rules, and preferences have to be learnt by autonomous agents while interacting with the environment and with human agents, who provide examples and indications grounded on their experience, in a process that, starting from simple and situated principles, allows the gradual acquisition of more complex rules. Following a developmental approach, we examine below how this process unfolds in humans, and in particular in children, who gradually align with social norms while exploring and learning in the environment, in order to explore how it can be implemented for AAA alignment.

2. *Autonomy and the acquisition of social norms in children*

Human autonomy can be defined as the capacity to think, decide, and act freely and independently on the basis of one's own reasoning and choices (Gillon, 1985). This capacity is not innate in a fully developed form: it emerges progressively from early infancy and requires continuous development throughout childhood. A substantial body of research highlights the importance of autonomy as a fundamental component of children's physical and psychological development because it would have a positive impact on mental health, self-esteem, and academic achievement (Oliverio Ferraris, 2024). Providing children with opportunities to explore and test their abilities, such as making decisions and managing problem-solving, facilitates the acquisition of essential competencies that will prove beneficial in adulthood. In this respect, autonomy plays a key role in fostering interpersonal skills, including communication abilities, relational competence, and the capacity to understand and apply social norms. Importantly, becoming autonomous does not merely entail the ability to perform tasks independently, but it involves the development of deeper competencies related to the appropriate management of diverse situations and social relationships.

The development of autonomy in children is a gradual process closely linked to the acquisition of rules, basically beginning shortly after weaning. In the early stages of life, caregivers act as primary agents in promoting autonomy and serve as behavioural models through imitation: as Tomasello (2019) argues, imitation reflects a tendency toward conformity with adult models, through which children acquire behavioural norms, that is, an understanding of what is appropriate or inappropriate in specific contexts, thus allowing their integration into a social group. Autonomy is therefore closely intertwined with the acquisition of social norms, understood as «often informal rules that structure human behaviour, regulating what is appropriate, required, prohibited or permitted» (Kelly, 2020, p. 36), since both contribute to shaping individual behaviour in relation to others. Like autonomy, the learning of social norms is gradual (Tomasello, 2025): around the age of three, children begin to understand and autonomously apply them, frequently employing generic normative language in contexts involving third-party intervention, and by the age of five they become capable of generating norms themselves. The emergence of autonomy is also closely tied to the development of self-regulation, shown to be a central mediator between environmental experience and social competence (Barkley, 2012). As children learn to regulate their behaviour, they gradually acquire the ability to internalize and apply shared norms. However, alignment with social norms is not simply the result of external imposition, but depends on the individual's autonomous capacity to interpret, contextualize, and flexibly apply them. From this perspective, autonomy constitutes a necessary condition for the adaptive management of social norms: in its absence, individuals would be limited to following rules mechanically, without the ability to adjust them to varying circumstances. Conversely, excessive or unregulated autonomy may result in deviations from socially accepted expectations. Only through exploration and participation in collective activities, therefore, do children progressively internalize norms, thus contributing to their maintenance as mechanisms of social coordination. It is also interesting to note that the acquisition and transmission of social norms is not unique to humans: research shows that it belongs to primates as well (Andrews et al., 2024).

If, in humans, and especially in childhood, autonomy serves as the foundation for acquiring essential competencies for navigating the social world, it is worth questioning if it fulfills a comparable function in AAA. Furthermore, one should consider whether the tendency towards alignment observed in humans constitutes a naturally selected disposition, and therefore one that cannot be assumed as given in artificial agents. Building on the developmental distinction between fixed-rule execution and genuine norm internalisation in children, the key question is how AAA can learn and internalise norms in a way that supports flexible, context-sensitive, and robust alignment in novel situations.

3. *A developmental model of norm alignment in autonomous agents*

The key insight of our proposal is that the acquisition, internalisation, and alignment with social norms, in both humans and AAA require a learning process that is (i) embodied and situated, (ii) intrinsically motivated, (iii) socially interactive, and (iv) structured into developmental stages. Embodiment and situatedness are necessary because norms are constitutively tied to the contexts in which they are enacted: they cannot be learned as abstract propositions but must be acquired through participation. Intrinsic motivations are necessary because the internalisation of norms requires an agent that is genuinely invested in understanding its environment and that can sustain exploration in novel and unpredictable situations. Social interaction is necessary because norms are inherently intersubjective: they require agents to be sensitive to others' reactions, adapt to their expectations, and adjust their behaviour accordingly. Developmental stages are necessary because norms are organised hierarchically: basic norms must be internalised before complex ones, and the capacity to manage normative conflict requires a reflective competence that can only be developed on the basis of prior practical mastery. Following Dennett (1996), the status of moral agent is not innate but attributed gradually on the basis of the capacity to manage degrees of freedom responsibly: as in humans the freedom to exercise autonomy is granted progressively alongside the recognition of moral status, our model similarly proposes that the autonomy of AAA should be extended in proportion to their ability to manage it in alignment with shared norms. This has several concrete design implications. First, AAA should be trained in staged environments that gradually increase in normative complexity, beginning with simple, clear, highly scaffolded normative contexts and then progressing toward complex, unstructured, and conflictual contexts. Second, AAA's access to degrees of freedom should be made contingent on their demonstrated alignment: AAA should earn their autonomy through performance, not receive it upfront. Third, the training process should include human interactions that provide normative feedback in a rich and contextualised way.

It is worth emphasizing that the proposed model raises a number of issues and implications, including the nature and limits of AAA's moral agency, governance, and accountability. On whether functional moral agents, that is, systems that behave as if guided by moral principles, are genuinely moral agents or merely sophisticated simulacra (Floridi & Sanders, 2004), we argue that, even if AAA can exhibit increasingly sophisticated norm-management capacities, this does not make them morally responsible, since moral responsibility, as traditionally understood, requires free will and intentionality in a sense that cannot be attributed to actual or foreseeable systems. A more relevant issue raised by our model concerns the distribution of normative authority and responsibility between humans and AAA. If the goal is to produce genuinely norm-aligned agents, the responsibility for such alignment cannot fall solely on their designers, who cannot anticipate all the normative contexts in which they will operate.

Since AAA must be able to interpret norms, identify gaps and conflicts, and exercise judgement in their application, some degree of normative authority must be delegated to them, but only gradually, as they demonstrate reliable, flexible, and context-sensitive competence in managing social norms, including the ability to justify normative choices, respond to correction, and integrate feedback into subsequent behaviour. In the absence of such capacities, expanding autonomy may produce systems whose actions become difficult to predict, govern, or align with evolving social norms. In child development, this risk is mitigated through continuous social feedback and pedagogical scaffolding. A similar logic should guide the governance of AAA: autonomy should be expanded only in parallel with mechanisms of ongoing normative feedback and correction, both during training and throughout operational deployment. In this regard, the AI regulatory sandboxes introduced by the AI Act may provide a suitable framework for simulating and testing progressive forms of social interaction, normative learning and alignment under controlled supervisory conditions.

The goal is not to produce perfectly aligned AAA through training alone, but to create systems whose alignment is maintained dynamically through their ongoing participation in normative social life. This robust and flexible normative competence enables genuine co-existence and collaboration between human and artificial agents in a shared social world. If autonomy is the condition of possibility for the internalisation of norms in children, then cultivating appropriate forms of autonomy in AAAs is not a threat to alignment, but rather its precondition.

**Acknowledgments**
This work was partially funded by the European Union's Horizon 2020, research and innovation programme under GA 101070381 ('PILLAR-Robots - Purposeful Intrinsically-motivated Lifelong Learning Autonomous Robots').